\documentclass[journal]{IEEEtran} 
\usepackage{algpseudocode}
\usepackage{algorithm}
\usepackage{cite}
\usepackage{amsmath,amssymb,amsfonts}
\usepackage{graphicx}
\usepackage{textcomp}
\usepackage{xcolor}
\usepackage{multirow}
\usepackage{caption}
\usepackage{subcaption}
\usepackage{titlesec}
\usepackage{gensymb}
\usepackage{color,soul}
\usepackage{array}
\usepackage{setspace}
\usepackage{caption}
\usepackage{booktabs}
\usepackage{multirow}
\usepackage{titlesec}
\usepackage{makecell}
\usepackage{bbm}
\usepackage{arydshln}
\usepackage[colorlinks=true, linkcolor=blue, urlcolor=blue, citecolor=green]{hyperref}
\usepackage{xspace}
\usepackage{caption}           
\newcommand{\method}{\textsc{Frieren}\xspace} 

\def\BibTeX{{\rm B\kern-.05em{\sc i\kern-.025em b}\kern-.08em
    T\kern-.1667em\lower.7ex\hbox{E}\kern-.125emX}}
\begin{document}
\setcounter{secnumdepth}{4}
\title{FRIEREN: Federated Learning with Vision-Language Regularization for  Segmentation}
\author{%
  \IEEEauthorblockN{Ding-Ruei Shen\IEEEauthorrefmark{1}\IEEEauthorrefmark{2}}\\
  \IEEEauthorblockA{\IEEEauthorrefmark{1}Eindhoven University of Technology, Eindhoven, The Netherlands}\\
  \IEEEauthorblockA{\IEEEauthorrefmark{2}NXP Semiconductors, Eindhoven, The Netherlands}%
}
\titlespacing{\section}{0pt}{*2}{*0}
\titlespacing{\subsection}{0pt}{*3.0}{*0}
\raggedbottom
\maketitle
\begin{abstract}
\textit{Federeated Learning~(FL) offers a privacy-preserving solution for Semantic Segmentation~(SS) tasks to adapt to new domains, but faces significant challenges from these domain shifts, particularly when client data is unlabeled. However, most existing FL methods unrealistically assume access to labeled data on remote clients or fail to leverage the power of modern Vision Foundation Models~(VFMs). Here, we propose a novel and challenging task, FFREEDG, in which a model is pretrained on a server's labeled source dataset and subsequently trained across clients using only their unlabeled data, without ever re-accessing the source. To solve FFREEDG, we propose \method, a framework that leverages the knowledge of a VFM by integrating vision and language modalities. Our approach employs a Vision-Language decoder guided by CLIP-based text embeddings to improve semantic disambiguation and uses a weak-to-strong consistency learning strategy for robust local training on pseudo-labels. Our experiments on synthetic-to-real and clear-to-adverse-weather benchmarks demonstrate that our framework effectively tackles this new task, achieving competitive performance against established domain generalization and adaptation methods and setting a strong baseline for future research.}
\end{abstract}

\section{Introduction}\label{sec:intro}
The transformation of raw visual data, captured by cameras and other imaging sensors, into actionable intelligence is a fundamental objective in numerous technological domains. Semantic segmentation, a computer vision task that assigns a semantic label to every pixel in an image, plays a pivotal role by providing dense scene understanding for decision-making across applications such as autonomous driving and medical imaging \cite{long2015fully,cordts2016cityscapes,ronneberger2015u}. This detailed representation enables self-driving vehicles to separate roads, pedestrians, and obstacles for safe navigation and allows clinicians to delineate anatomical structures and abnormalities in scans to improve diagnostics and downstream workflow automation \cite{cordts2016cityscapes,ronneberger2015u}.
\vspace{0.5\baselineskip}

Recent years have witnessed the rise of Vision Foundation Models (VFMs), e.g., CLIP~\cite{radford2021learning}, Segment Anything~\cite{kirillov2023segment}, and DINOv2~\cite{oquab2023dinov2}, that exhibit strong generalization across diverse vision tasks after large-scale pretraining. Their transferable visual representations can substantially reduce task-specific annotation needs when adapted to downstream segmentation. Critically for semi-supervised segmentation, Vision-Language models~(VLMs) inject rich semantic priors that help disambiguate visually similar categories. For instance, SemiVL~\cite{hoyer2024semivl} demonstrates that integrating VLM guidance with consistency training improves mean Intersection over Union~(mIoU) and reduces confusions among classes with similar appearance under scarce labels. Related language-conditioned segmentation also reports benefits for ambiguous urban classes such as person vs.\ rider, supporting the intuition that language cues are relatively robust under domain shift. LC-MSM~\cite{kim2024lc} further advances this by proposing a language-conditioned masked segmentation model that jointly learns context relations and domain-agnostic information.

\begin{figure}[t]     
  \centering
  \includegraphics[scale=0.45]{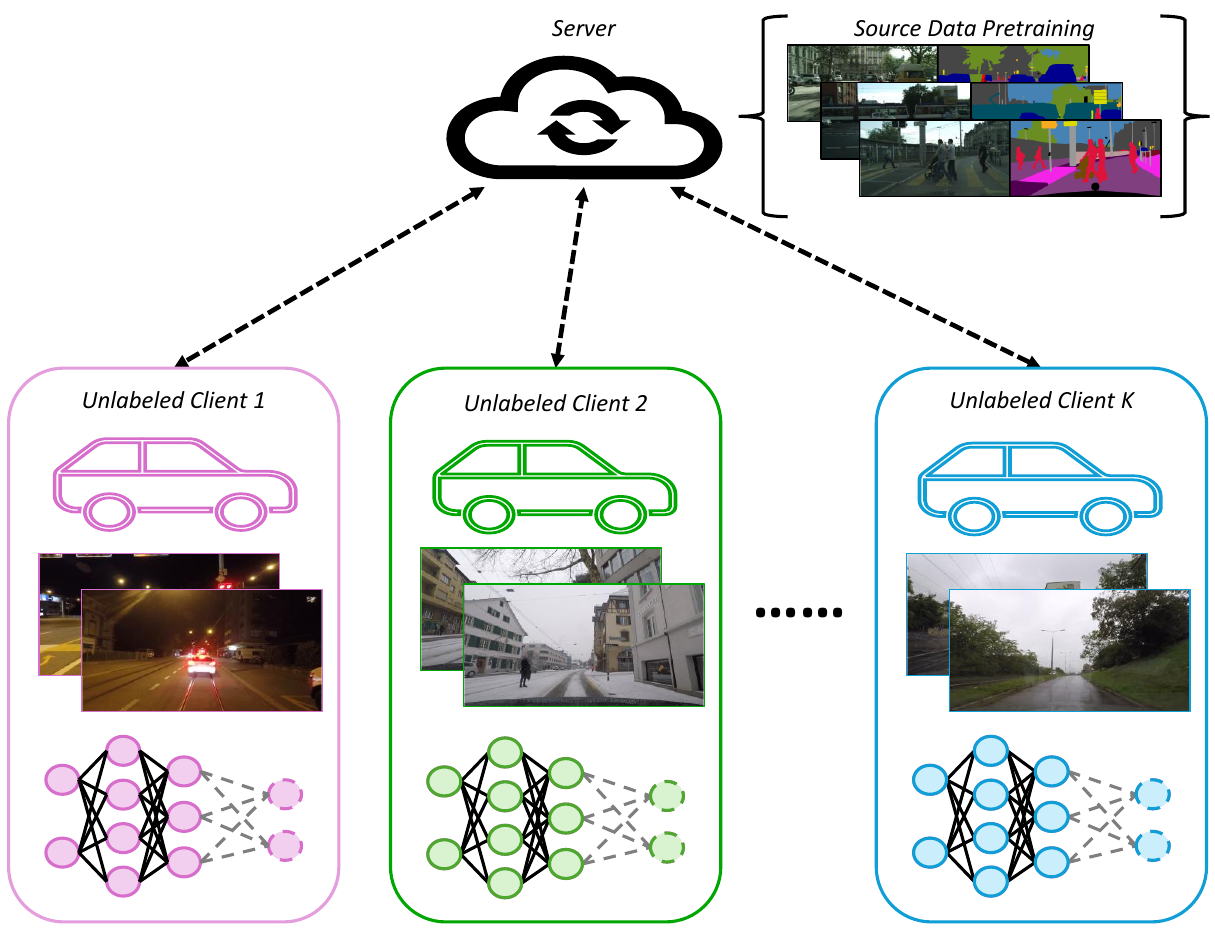}  
  \caption{Federated source-Free Domain Generalization (FFREEDG) framework. A central server coordinates the training of a global model with multiple clients. Each client has access only to its own private, unlabeled data, which comes from a distinct domain, such as different weather or lighting conditions. The clients perform local, unsupervised training before the server aggregates their updates to create a single model that can generalize across these diverse environments.}
  \label{fig:fig1}
\end{figure}

\vspace{0.5\baselineskip}
Although VFMs can lower the labeled-data burden, many practical pipelines still centralize data from multiple sites to a single server for pretraining and fine-tuning, which raises significant privacy concerns in domains such as healthcare and autonomous vehicles~\cite{kairouz2021advances}. The need to protect sensitive data in these fields necessitates the exploration of decentralized learning approaches to mitigate the risks associated with centralizing large, private datasets. Federated Learning (FL)~\cite{kairouz2021advances} has emerged as a promising solution, enabling collaborative model training across multiple clients~(e.g., hospitals, autonomous vehicles)~without the need to share raw data directly, thereby preserving privacy. This is particularly valuable in fields like healthcare and autonomous vehicles where data privacy is paramount~\cite{sheller2020federated,chellapandi2023federated,shenaj2023learning}.
\vspace{0.5\baselineskip}

Most existing Federated Learning approaches to semantic segmentation rely on relatively old architectures without fully exploiting the advantages of VFMs~\cite{shenaj2023learning,alphonse2025federated,mu2023fedproc,rizzoli2025cars,shiri2023multi,nalawade2021federated,tolle2025real}. To address domain gaps in federated learning between pretrained datasets and client datasets, some utilize domain adaptation techniques and assume clients have specific styles \cite{shenaj2023learning,rizzoli2025cars,fantauzzo2022feddrive,lian2025feds2r}, while others focus on supervised learning without considering unlabeled datasets or scarce-labeled-dataset cases~\cite{liu2024fedfms}. For segmentation tasks, many semi-supervised federated learning approaches focus on medical imaging applications \cite{zheng2024federated,ma2024model,wang2024feddus}, where the realm of urban scene datasets for segmentation remains underexplored. Encouragingly, first steps toward federated foundation models in segmentation are emerging:~FedFMS~\cite{liu2024fedfms} adapt~SAM~\cite{kirillov2023segment} with communication-efficient adapters in FL for medical images, showing that FM priors can be exploited under federation; Yet, semi-/unsupervised federated segmentation remains mostly studied in medical imaging, leaving urban scene parsing comparatively underexplored.
\vspace{0.5\baselineskip}

Inspired by Source-Free Domain Adaptation for Semantic Segmentation~\cite{shenaj2023learning}, we consider a more realistic setting named \textbf{F}ederated source-\textbf{Free} \textbf{D}omain \textbf{G}eneralization (FFREEDG) (\textcolor{green}{Fig.~\ref{fig:fig1}}). Compared to Federated source-Free Domain Adaptation~(FFREEDA)~\cite{shenaj2023learning}, our work does not apply domain adaptation techniques but utilizes foundation models like CLIP to achieve strong generalization. FFREEDG builds on FFREEDA, which in turn adopts the Source-Free Domain Adaptation~(SFDA)~setting~\cite{liu2021source}. Under SFDA, the source dataset is not re-accessed during adaptation because, in applications such as autonomous driving, the source data are private or proprietary; thus only the released source model and unlabeled target data are available. Our scenario simulates the situation where the server-pretrained model has zero knowledge about client styles or priors in advance. Therefore, style-driven domain adaptation techniques are not possible in our case, and to further simulate real-world cases where labeled data is expensive, the model is only able to use source-labeled data on the server for pretraining. After pretraining, the model cannot access source data, so the local training on clients is fully unsupervised. 
\vspace{0.5\baselineskip}

Although we do not exploit domain adaptation techniques, inspired by open segmentation and language-conditioned models such as LC-MSM, SemiVL, and SemiDevil~\cite{basak2025semidavil}, we integrate text embeddings with visual features to inject domain-agnostic priors; because the source and target domains share the same class vocabulary, the language side remains invariant under domain shift~\cite{kim2024lc}. VLM models like CLIP are pretrained on large web-scale image–caption datasets, and their ability to link semantics and images has been shown in many works~\cite{radford2021learning}. They are particularly effective at discerning ambiguous classes, which is very useful in datasets like Cityscapes~\cite{cordts2016cityscapes}, where some classes are difficult to distinguish, such as rider vs.\ person and car vs.\ truck~\cite{kim2024lc}.
\vspace{0.5\baselineskip}

The semi-supervised techniques we apply originate from UniMatch~V1~\cite{yang2023revisiting} and UniMatch~V2~\cite{yang2025unimatch}, which adapt FixMatch-style~\cite{sohn2020fixmatch} learning strategies in which the model’s predictive power is based on labeled data and pseudo-labels from weak-to-strong augmentation, applying a consistency loss to enhance feature alignment. This training framework, after modification, can also be applied to unsupervised self-training. Our experiments on centralized learning indicate that, in unsupervised cases, performance can slightly increase, while in semi-supervised cases, performance is largely improved as well. Based on this observation, it is suitable for FFREEDG. Together, our model is able to handle semi-supervised and unsupervised scenarios with integrated learning strategies using visual and text modalities. To the best of our knowledge, no previous work in federated learning provides such a framework.
\vspace{0.5\baselineskip}

In summary, this paper proposes \method{}~(\textbf{F}ederated \textbf{R}easoning with V\textbf{i}sion–languag\textbf{e} \textbf{Re}gularization for Segmentatio\textbf{n}) to tackles the semi- and unsupervised federated learning problems by building a Vision-Language fusion model using foundation models like CLIP, addressing privacy concerns while ensuring the performance of distributed learning. Our experiments conducted on two heterogeneous datasets constructed from ACDC~\cite{sakaridis2021acdc} and Cityscapes indicate the efficiency of this framework. Our contributions are as follows:
\begin{enumerate}
\item We formulate FFREEDG, a source-free federated generalization framework for semantic segmentation that eliminates the need for style-specific adaptation while preserving privacy.
\item We propose a Vision-Language-guided semi-/unsupervised federated framework that fuses CLIP text embeddings with visual features and uses class prompts to reduce semantic ambiguities under label scarcity, extending UniMatch-style guidance to the federated setting.
\item We conduct an empirical study on ACDC and Cityscapes under federated semi-supervised and fully unsupervised training, demonstrating that language guidance complements pseudo-label consistency and that, in semi-supervised settings, performance approaches that of fully supervised FL while substantially reducing annotation requirements.
\end{enumerate}


\section{Related Work}\label{sec:relate}
This section reviews prior work relevant to our setting. Section~\ref{sec:da} revisits domain adaptation (DA) and its assumptions and limitations. Section~\ref{sec:dg} surveys domain generalization (DG) and clarifies its connection to our FFREEDG formulation. Section~\ref{sec:sss} relates our approach to semi-supervised semantic segmentation. Sections~\ref{sec:flv} and~\ref{sec:ffm} summarize federated learning for vision and federated foundation models, respectively. Finally, Section~\ref{sec:srg} synthesizes these threads and highlights the remaining research gap.

\subsection{Domain Adaptation (DA)}\label{sec:da}
Domain Adaptation narrows the gap between a labeled source domain and an unlabeled target domain. Typical approaches for DA include: (i) explicit domain divergence matching in feature space~\cite{long2015learning, tzeng2019deep}, (ii) adversarial alignment with a domain discriminator~\cite{luo2019taking,tsai2018learning,michieli2020adversarial}, (iii) image-to-image translation to render source images in the target style~\cite{hoffman2018cycada,pizzati2020domain,toldo2020unsupervised}, plus efficient, non-trainable variants such as Fourier Domain Adaptation (FDA)~\cite{yang2020fda}, and (iv) self-training with confidence-aware pseudo labels~\cite{mei2020instance,zou2018unsupervised}. 
These approaches form the standard toolkit for source-to-target adaptation in segmentation. However, conventional DA methods typically assume access to the labeled source data during adaptation, which is often infeasible due to privacy, copyright, or resource constraints~\cite{liu2021source}. Source-Free~Domain~Adaptation~(SFDA)~\cite{li2024comprehensive} relaxes this assumption. SFDA methods adapt a model to the target domain using only a pretrained source model and the unlabeled target data, with no source images available.

\subsection{Domain Generalization for Semantic Segmentation}\label{sec:dg}
Domain~generalization~(DG) trains models on one or more source domains to perform robustly on unseen target domains without accessing any target data during training~\cite{rafi2024domain}. This is particularly crucial for semantic segmentation in autonomous driving, where domain shifts arise from variations in weather, lighting, and geography, potentially degrading model performance. A key category in DG is~\emph{data manipulation}, which augments source data to simulate diverse “pseudo-domains” and encourages learning of domain-invariant features. As outlined in the DG survey by~\textit{Rafi~et~al.}~\cite{rafi2024domain}, these approaches include data augmentation, domain randomization, data generation, and adversarial learning.
\vspace{0.5\baselineskip}

Within data augmentation, Fourier-based methods such as Amplitude Mix~\cite{xu2021fourier}~perturb the amplitude spectrum while preserving phase information, altering image appearance without disrupting semantic structure. Domain-randomization techniques include texture-level strategies like Global/Local Texture Randomization~\cite{peng2021global}, which incorporate diverse textures from paintings, and FSDR~\cite{huang2021fsdr}, which identifies and randomizes domain-variant frequency components while retaining the semantic structures of images and domain-invariant features.
\vspace{0.5\baselineskip}

In federated DG settings, methods like FedS2R~\cite{lian2025feds2r} identify unstable classes through cross-client prediction inconsistencies on an unlabeled server set, then synthesize targeted images using diffusion models guided by scene-aware prompts, followed by global model distillation. However, FedS2R relies on indirect client knowledge—specifically, model outputs from clients—on a server dataset drawn from the target (real-world) domain. This assumes access to target-like unlabeled data and prior knowledge of client styles through their models, which is incompatible with our FFREEDG setting. In contrast, FFREEDG operates under stricter assumptions: the server is trained solely on source-labeled data, without access to client domains or target-like data, and clients perform fully unsupervised local training with no data sharing or server-side augmentation.
\vspace{0.5\baselineskip}

Recently, vision–language and prompt-based DG methods have gained traction, leveraging foundation models like CLIP to exploit semantic priors from text embeddings, which are robust to domain shifts~\cite{radford2021learning}. For instance, VLTSeg~\cite{hummer2023vltseg} employs full-parameter fine-tuning of vision–language models to achieve state-of-the-art~(SOTA) performance in semantic segmentation tasks. Similarly, Rein~\cite{wei2024stronger} introduces lightweight adapters to fine-tune backbone models, balancing efficiency and performance. These methods demonstrate that fine-tuning foundation models—without relying on complex style augmentation or randomization—can yield competitive or superior DG results for segmentation. Inspired by VLTSeg and Rein, our framework avoids traditional style augmentation or randomization strategies and instead leverages fine-tuning of vision–language foundation models, such as CLIP, to achieve high performance in semantic segmentation.

\subsection{Semi-Supervised Learning for Segmentation}\label{sec:sss}
Semi-supervised semantic segmentation~(SSS) aims to alleviate the annotation burden of pixel-wise labeling by leveraging a small set of labeled images together with a larger unlabeled dataset. The key challenge lies in extracting reliable supervisory signals from the unlabeled data to guide learning and improve generalization.
\vspace{0.5\baselineskip}

As outlined in a recent survey~\cite{pelaez2023survey}, SSS methods can be classified into five main categories: adversarial methods, pseudo-labeling, consistency regularization, contrastive learning, and hybrid approaches. Adversarial methods, often using generative adversarial networks~(GANs), encourage predictions indistinguishable from ground truth~(e.g., generative~\cite{li2021semantic} and non-generative~\cite{ke2020guided}). Contrastive learning pulls similar features closer while pushing dissimilar ones apart~\cite{alonso2021semi,liu2021bootstrapping}. Hybrid methods combine multiple categories for enhanced performance~\cite{zou2020pseudoseg,ke2022three}. However, current solutions are primarily bifurcated into two dominant approaches: consistency regularization~\cite{french2019semi,li2020semi,kim2001structured} and pseudo-labeling/self-training~\cite{yang2022st++,teh2022gist}.
\vspace{0.5\baselineskip}
A representative hybrid method is FixMatch~\cite{sohn2020fixmatch}, an image classification approach that filters pseudo-labels from weakly augmented views and applies them to strongly augmented counterparts, combining pseudo-labeling with consistency principles. Building on these foundations, UniMatch~V1~\cite{yang2023revisiting} adapts FixMatch to segmentation by integrating image- and feature-level perturbations into a unified weak-to-strong pipeline. UniMatch~V2~\cite{yang2025unimatch} improves upon this by incorporating complementary dropout and an exponential~moving~average~(EMA)~model, achieving improved training stability and SOTA results. In a different direction, SemiVL~\cite{hoyer2024semivl} tackles semantic ambiguity in low-label regimes by integrating CLIP-based vision–language guidance. Through spatial fine-tuning and a language-guided decoder, it enhances boundary accuracy and class discrimination.
\vspace{0.5\baselineskip}

Our work builds upon these advances in two key ways. First, we extend SemiVL’s vision–language decoder to inject strong semantic priors derived from CLIP, enhancing disambiguation in label-scarce settings. Second, we adopt the training strategies of UniMatch~V2—complementary dropout and EMA—to strengthen consistency learning. Importantly, we adapt these components to our FFREEDG setting, which imposes stricter constraints: training must proceed in a source-free, federated, and unsupervised manner, without access to client domain priors or target data.

\subsection{Federated Learning for Vision Tasks}\label{sec:flv}
Federated learning~(FL) enables collaborative model training across multiple clients without sharing raw data, making it especially appealing for privacy-sensitive applications such as medical imaging or autonomous driving. In the context of semantic segmentation, recent methods have explored FL under both supervised and source-free conditions. For example, LADD~\cite{shenaj2023learning}~clusters clients by image style and combines global and cluster-specific aggregation for source-free domain adaptation. HyperFLAW~\cite{rizzoli2025cars} extends FL to multi-viewpoint~(e.g., car/drone) and adverse-weather settings using weather-aware batch normalization and hyperbolic prototype alignment. FedS2R~\cite{lian2025feds2r} addresses synthetic-to-real segmentation under a one-shot FL setup, leveraging inconsistency-driven data augmentation and multi-client knowledge distillation. Meanwhile, HSSF~\cite{ma2024model} investigates model-heterogeneous, semi-supervised FL with self-assessment and reliable pseudo-label generation modules. While these works address domain heterogeneity or data decentralization, most assume labeled data is available locally or neglect the power of modern foundation models to aid generalization.

\subsection{Federated Foundation Models}\label{sec:ffm}
Building on the success of large-scale pretraining, recent work has begun exploring Federated Foundation Models~(FedFMs~\cite{ren2025advances})—a paradigm that merges pretrained models with FL to tackle data heterogeneity, scarcity, and communication bottlenecks. Foundation models like CLIP and SAM can serve as strong priors, while client-side fine-tuning supports privacy preservation.
\vspace{0.5\baselineskip}

FST-CBDG~\cite{yan2024lightweight}~uses a frozen CLIP image encoder and class text embeddings for unsupervised federated classification, training a lightweight local classifier via pseudo-label self-training. In segmentation, FedFMS~\cite{liu2024fedfms}~applies federated fine-tuning of SAM for medical image segmentation, either via full-model~aggregation~(FedSAM) or a lightweight~adapter~(FedMSA), achieving strong results with reduced communication costs. FedCLIP~\cite{lu2023fedclip}~introduces adapter-based tuning of CLIP in FL, improving generalization while keeping communication efficient. Similarly, CLIP2FL~\cite{shi2024clip} combines client-side CLIP distillation with server-side prototype-guided aggregation, showing improved robustness under data imbalance and heterogeneity. These studies illustrate that foundation models can significantly enhance FL performance.
\vspace{0.5\baselineskip}

However, existing FedFM approaches either assume access to labeled client data~(e.g., FedFMS), focus on classification rather than dense prediction~(e.g., FST-CBDG), or rely on indirect prior knowledge from clients’ model predictions~(e.g., FedS2R). In contrast, our work addresses a more challenging setting:~unsupervised federated segmentation without any client-side labels. We build on a single shared vision foundation model~(CLIP) as the backbone and enable federated training of a semantic segmentation model without access to target labeled data, domain priors. This is, to our knowledge, the first effort to combine foundation models with source-free, unsupervised federated segmentation, filling a gap in the current literature.

\subsection{Summary and Research Gap}\label{sec:srg}
Across the literature, prior work tackles isolated challenges: DA address domain shift using target data, which may be inaccessible; SSS reduces label cost but assumes centralized data; and FL preserves privacy but lacks robustness to shift or label efficiency. In contrast, our framework unifies these concerns: we integrate vision–language pretraining, unsupervised learning, and federated training—\emph{all under a source-free, target-label-free regime}. This enables generalization to unseen domains without relying on target~labeled~labels, domain~priors, or re-accessing centralized data.

\section{Problem Setting}

We study \emph{Federated, source-Free, domain Generalization for semantic segmentation}~(FFREEDG). A central coordinator communicates with~$K$~clients indexed by~$\mathcal{K} \!=\! \{1,\dots,K\}$. Let~$\mathcal{X}$~denote the image space, $\mathcal{Y} \!=\! \{1,\dots,C\}$~the label set~(shared across all domains), and~$N_p\!=\!H\!\times\!W\times\!C$~the number of pixels per image. The segmentation network is \mbox{$f(x;w):\mathcal{X}\!\to\!\mathbb{R}^{N_p\times C}$} with parameters~$w$, producing per-pixel class posteriors.

\subsection{Server Pretraining}
The server owns a labeled source dataset \mbox{$\mathcal{D}_{\mathrm{src}}=\{(x_i^{\mathrm{s}},y_i^{\mathrm{s}})\}_{i=1}^{n_{\mathrm{s}}}$} drawn from a distribution \mbox{$P_{\mathrm{s}}(x,y)$}. An initialization~$w_0$~is learned by supervised training and the source data are then discarded and never used during federation:
\begin{equation}
  w_0 \in \arg\min_{w}\;
  \mathbb{E}_{(x,y)\sim P_{\mathrm{s}}}\big[\ell_{\mathrm{sup}}(f(x;w),y)\big].
  \label{eq:pretrain}
\end{equation}
This “source-free after pretraining’’ protocol follows prior FL work~\cite{liu2021source}~with server-side labeled sources.

\subsection{Clients Generalization}
Each client \mbox{$k\!\in\!\mathcal{K}$} holds an unlabeled local dataset 
\mbox{$\mathcal{D}_k=\{x^{(k)}_j\}_{j=1}^{n_k}$}, sampled from a client-specific 
image distribution \mbox{$P_k(x)$}. We assume~$P_k \in \Pi$, where~$\Pi$ is a meta-distribution over domains~(e.g., cities, sensors, weather); two clients may share the same latent domain, i.e.,~$P_k=P_h$ for some~$k\neq h$. Datasets are disjoint and non-IID; denote the total target pool~$\mathcal{D}_{\mathrm{tar}}=\bigcup_{k}\mathcal{D}_k$ with size~$n_{\mathrm{tar}}=\sum_k n_k$. During evaluation, the global model is tested on an unseen domain~$P_{\mathrm{tar}} \in \Pi$~that is not necessarily equal to any~$P_k$~(domain \emph{generalization}, not adaptation).

\subsection{Federated Protocol}
Training proceeds in communication rounds~$t=0,\dots,T\!-\!1$ with $w^{(0)}\!=\!w_0$.
At round~$t$, the server samples a subset~$\mathcal{S}_t\!\subseteq\!\mathcal{K}$, broadcasts~$w^{(t)}$, and each client~$k\!\in\!\mathcal{S}_t$ performs~$E$~local steps on an~\emph{unsupervised/semi-supervised}~surrogate objective~$\mathcal{L}_k$~(defined in the next section), yielding an updated~$w^{(t)}_k$.
The server aggregates the returned models into the next global iterate via a weighted rule~(FedAvg~\cite{mcmahan2017communication} shown here, other aggregators are possible):
\begin{equation}
  w^{(t+1)} \;=\; \sum_{k\in \mathcal{S}_t} \alpha_k^{(t)}\, w^{(t)}_k,
  \qquad
  \alpha_k^{(t)} \;=\; \frac{n_k}{\sum_{j\in \mathcal{S}_t} n_j}
  \label{eq:fedavg}
\end{equation}

\subsection{Learning objective}
Since client labels are absent, each site minimizes a surrogate loss~$\mathcal{L}_k$~that exploits unlabeled data~self-training~and optional side priors~(vision–language signals):
\begin{equation}
  w^\star \in \arg\min_{w}\; \sum_{k=1}^{K} \frac{n_k}{n_{\mathrm{tar}}}\; \mathcal{L}_k(w)
  \label{eq:global_obj}
\end{equation}
where, in our framework, $\mathcal{L}_k$~will combine (i) weak-to-strong consistency on local client dataset~$\mathcal{D}_k$, (ii) vision–language guidance, and (iii) distillation from fixed models (e.g., CLIP); we formalize these in Section~\ref{sec:Method}. This mirrors successful consistency designs for semi-supervised segmentation while remaining source-free during federation.

\subsection{Remarks}
\begin{itemize}
    \item Vision--language pretraining and guidance are based on the assumption that source and target domains share the same set of categories 
    ($\mathcal{Y} = \mathcal{Y}^{S} = \mathcal{Y}^{T}$).
    \item The Federated Aggregation equation~\eqref{eq:fedavg} is a placeholder; different aggregation algorithms can be plugged in without changing the system.
    \item If some clients provide a small labeled subset, a supervised term can be added to $\mathcal{L}_k$ with a scalar weight, recovering a semi-supervised federated regime.
\end{itemize}

\begin{figure*}[!t]                    
  \centering
  \includegraphics[width=\textwidth]{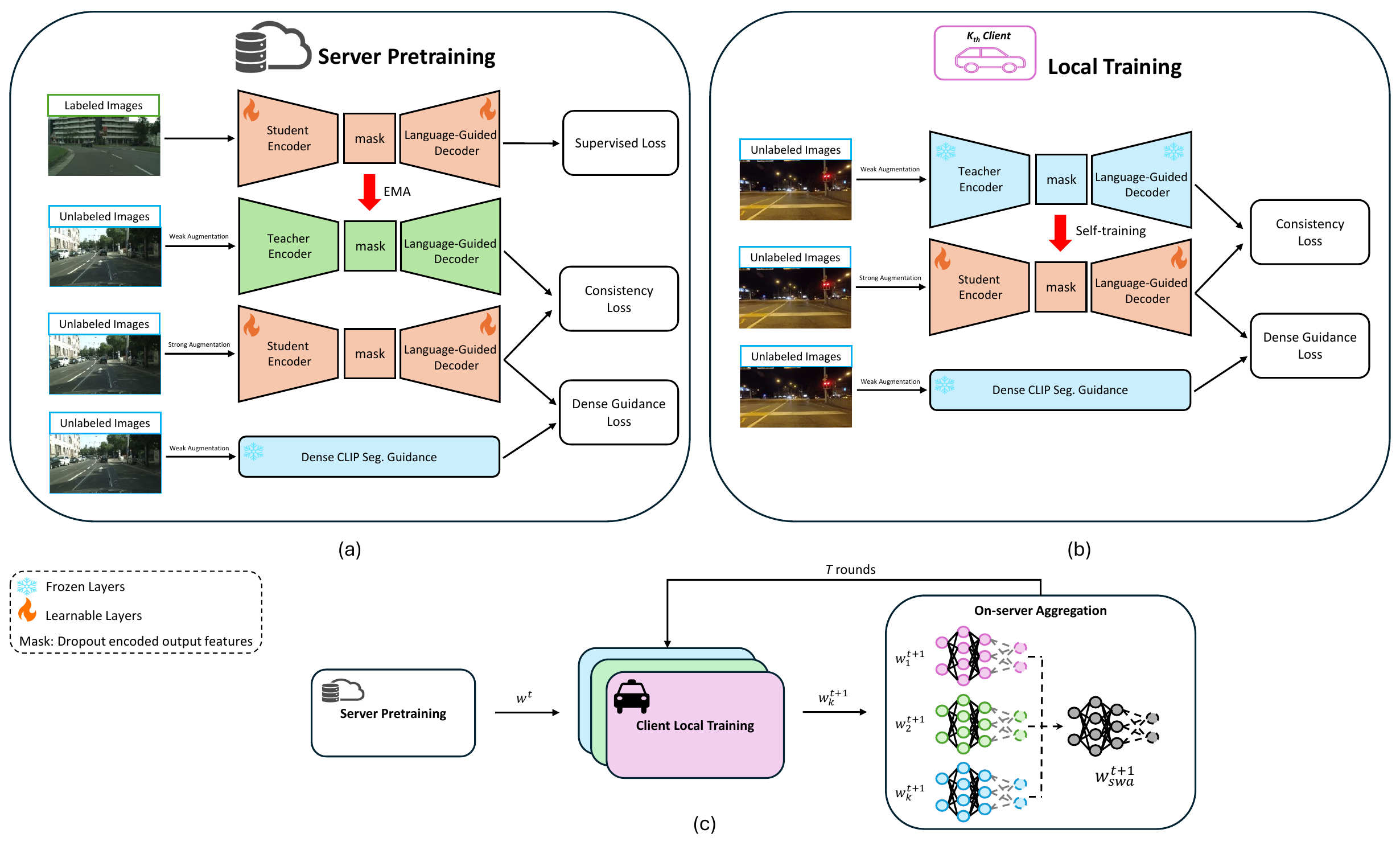}   
  \caption{Overview of FRIEREN. \textbf{Mask} denotes a dropout mechanism~(e.g., complementary dropout~\cite{yang2025unimatch}) that generates perturbed encoder features for consistency learning.
  \textbf{(a)}~Server pretraining:~a student–teacher setup with EMA updates on the teacher, supervised loss on labeled source images, weak–strong consistency on unlabeled images, and dense CLIP-based segmentation guidance via the language-guided decoder. Learnable modules and frozen modules are indicated in the figure. 
  \textbf{(b)}~Federated local training:~after pretraining, each selected client receives the global model and performs~(semi-)unsupervised self-training with a frozen teacher, consistency loss, and dense CLIP guidance; labeled client data~(if any) uses an additional supervised term. 
  \textbf{(c)}~Communication and aggregation:~the server broadcasts~$w^{t}$, collects client updates~$w_{k}^{t+1}$, and aggregates them; we use FedSWA~\cite{liu2025fedswa} for unsupervised clients and FedAvg~\cite{mcmahan2017communication} when supervision is available.}
  \label{fig:fig2}
\end{figure*}

\begin{figure*}[!t]
  \centering
  \includegraphics[width=\textwidth]{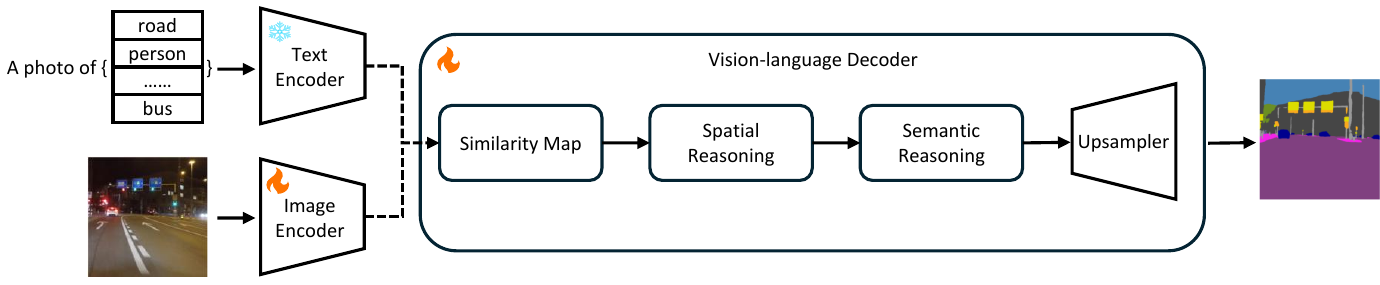}   
  \caption{Vision–language decoder in FRIEREN. FRIEREN adopts the language-guided decoder from SemiVL~\cite{hoyer2024semivl}. Class prompts are embedded by a frozen text encoder and paired with features from a trainable image encoder to form a dense similarity map. The decoder performs spatial reasoning followed by semantic reasoning, and a convolutional upsampler produces the final segmentation mask.}
  \label{fig:fig3}
\end{figure*}

\begin{algorithm}[t]
\caption{FRIEREN}
\label{alg:FRIEREN_alg}
\begin{algorithmic}[1]
\Require Labeled source set $\mathcal D_{\mathrm{src}}$ (server), client datasets $\{\mathcal D_k\}_{k=1}^K$, rounds $T$, local epochs $E$, aggregation operator $\textsc{Agg}\in\{\text{FedAvg}, \text{FedSWA}\}$
\State $w^{(0)} \gets \textsc{ServerPretrain}(\mathcal D_{\mathrm{src}})$ 
\For{$t = 0$ to $T-1$}
  \State Select participating clients $S_t \subseteq \{1,\ldots,K\}$ and broadcast $w^{(t)}$
  \ForAll{$k \in S_t$ \textbf{in parallel}}
     \State $w^{(t+1)}_k \gets \textsc{LocalTrain}(w^{(t)}, \mathcal D_k, E)$ 
  \EndFor
  \State $w^{(t+1)} \gets \textsc{Agg}\big(\{(w^{(t+1)}_k, n_k)\}_{k\in S_t}\big)$ 
\EndFor
\State \Return $w^{T}$
\end{algorithmic}
\end{algorithm}

\section{Method}\label{sec:Method}
In this section, we present \method, comprising two stages: centralized pretraining and federated learning. The procedure is summarized in Figure~\ref{fig:fig2} and \ref{fig:fig3}, and Algorithm~\ref{alg:FRIEREN_alg}. The centralized stage incorporates three key components—consistency regularization, a Vision-Language decoder, and dense CLIP distillation—designed to jointly strengthen semantic representation and robustness using labeled source data. This stage yields a well-initialized model that effectively leverages semantic cues from both vision and language modalities.
\vspace{0.5\baselineskip}

After centralized training, the model is transferred to a federated setting without accessing the original source data. In this federated stage, we address two scenarios: fully unsupervised and semi-supervised clients. For the semi-supervised case, where a small amount of labeled data is available at clients, we employ FedAvg to aggregate updates. In the unsupervised case, where only unlabeled data is available, we adopt FedSWA~\cite{liu2025fedswa} to stabilize optimization.

\subsection{Centralized Training Procedure}
\subsubsection{Consistency Regularization}
For a labeled minibatch~\((x, y)\)~and its prediction~\(\alpha_x\), an unlabeled minibatch~\(u\), draw weak and strong augmentations~\(\alpha_w, \alpha_s\). Let~\(p(\cdot \mid \cdot)\)~be the per-pixel softmax.
\begin{itemize}
\item \textbf{Supervised loss}:
\begin{equation}
\mathcal{L}_{\text{sup}} = \frac{1}{|\mathcal{B}_l| N_p} \sum_{(x,y)} \sum_{i=1}^{N_p} \log p(y_i \mid \alpha_w(x))
\end{equation}
This cross-entropy loss ensures direct supervision on labeled source data, grounding the model in accurate pixel classifications. In practice, we also use online~hard~example~mining~\cite{shrivastava2016training}, focusing on the hardest pixels to handle class imbalance.

\item \textbf{Weak-to-strong consistency}:\\
we first generate a \textbf{pseudo-label} 
for each pixel $i$ from the weakly augmented view:
\begin{equation}
\hat{y}_i = \arg\max_c \; p(c \mid \alpha_w(u))_i
\end{equation}

where $p(c \mid \alpha_w(u))_i$ is the per-class softmax probability at pixel $i$.
The confidence map is then defined as
\begin{equation}
c_i = \max_c \; p(c \mid \alpha_w(u))_i
\end{equation}
i.e., the maximum predicted probability over classes at that pixel.

We use this confidence map to decide whether the pseudo-label is reliable, via an indicator function:
\begin{equation}
\mathbbm{1}_i = \mathbbm{1}\!\big[ c_i \ge \tau \big]
\end{equation}
where~$\tau$~is a confidence threshold (e.g., $\tau=0.9$). 
Pixels with $c_i < \tau$ are ignored in training.

Finally, the weak-to-strong consistency loss is
\begin{equation}
\mathcal{L}_{\text{cons}}
= \frac{1}{|\mathcal{B}_u| N_p}
   \sum_{u} \sum_{i=1}^{N_p}
   \mathbbm{1}_i \,\log p(\hat{y}_i \mid \alpha_s(u))
\end{equation}
where the strong view~$\alpha_s(u)$~is supervised only on confident pseudo-labels. This regularization promotes invariance to augmentations and ensures that only confident pseudo-labels (where~$c_i \ge \tau$) are used for training, reducing the effect of noisy predictions.
Following UniMatch, we extend this by incorporating dual-stream augmentations: two strong views~\(\alpha_{s1}(u)\) and~\(\alpha_{s2}(u)\) are generated from the same weak view, and consistency is enforced bidirectionally:
\begin{equation}
\mathcal{L}_{\text{cons-dual}} = \mathcal{L}_{\text{cons}}(\alpha_{s1}, \alpha_{w}) + \mathcal{L}_{\text{cons}}(\alpha_{s2}, \alpha_{w}),
\end{equation}
where each term uses pseudo-labels from weak view to supervise two strong views. Feature-level dropout is applied post-encoder, dropping channels with probability~\(p_d = 0.5\), to simulate perturbations in the latent space. Additionally, we incorporate an exponential~moving~average~(EMA) update of the model parameters to stabilize training and obtain slightly improved performance. 
A teacher~model~$\theta^t$ is maintained as a smoothed version of the student: 
\begin{equation}
\theta^t \leftarrow \gamma \, \theta^t + (1-\gamma)\,\theta^s,
\end{equation}
where the momentum~$\gamma$ is set to~$0.996$. 
This moving average reduces the variance of pseudo-labels and provides a more reliable teacher signal during training.

\end{itemize}

The centralized training objective is:
\begin{equation}
\mathcal{L}_{\text{cent}} 
= \mathcal{L}_{\text{sup}} 
+ \lambda_{\text{cons}} \, \mathcal{L}_{\text{cons-dual}}
\end{equation}

\subsubsection{Unsupervised~Self-training}
In the unsupervised setting, no ground-truth labels are available. 
We therefore replace the supervised term with a teacher-guided self-training loss, 
where a frozen teacher~$\theta^t$ produces pseudo-label. For an unlabeled minibatch~$u$, let the teacher's per-pixel softmax be
$p^{t}(c \mid \alpha_w(u))_i$. We define the teacher pseudo-label, confidence
map, and confidence mask as follows:
\begin{equation}
\begin{aligned}
\tilde{y}_i       &= \arg\max_c\, p^{t}(c \mid \alpha_w(u))_i, \\[4pt]
c^{t}_i           &= \max_c\,    p^{t}(c \mid \alpha_w(u))_i, \\[4pt]
\mathbbm{1}^{t}_i &= \mathbbm{1}\!\big[ c^{t}_i \ge \tau_t \big].
\end{aligned}
\end{equation}

The unsupervised objective becomes
\begin{equation}
\mathcal{L}_{\text{unsup}}
= \lambda_{\text{t}} \, \mathcal{L}_{\text{t}}
+ \lambda_{\text{cons}} \, \mathcal{L}_{\text{cons-dual}}
\end{equation}
which is identical to the centralized objective except that~$\mathcal{L}_{\text{sup}}$ is replaced by the teacher-supervised~$\mathcal{L}_{\text{t}}$.
\vspace{0.5\baselineskip}

\subsubsection{Vision–Language Decoder}
See~Figure~\ref{fig:fig3}. We reuse the language-guided decoder from SemiVL to inject CLIP’s vision–language aligned features into the segmentation head. 
Given per-pixel~visual~features~$v_i \in \mathbb{R}^d$ from the CLIP vision encoder and class text embeddings~$t_c \in \mathbb{R}^d$ over N classes, we form a dense similarity tensor
\[
S_{i,c} \;=\; \big\langle \mathrm{norm}(v_i),\, \mathrm{norm}(t_c) \big\rangle,
\qquad S \in \mathbb{R}^{H \times W \times N},
\]
which acts as language guidance for decoding.

The decoder follows the~\emph{spatial–semantic decoupling} principle of~SemiVL.~(i)~\textbf{Spatial reasoning:}~each class channel of~$S$ is first refined independently and embedded into a similarity map and then inject multi-scale context~(a convolution layer followed by an ASPP~module~\cite{chen2018encoder}).  
(ii)~\textbf{Semantic reasoning:}~inter-class relations are modeled per pixel using a small transformer that attends between the refined pixel token and projected text tokens; this transfers class semantics without introducing spatial interactions at this stage. Finally, two transpose-convolution upsampling blocks with skip connections from the encoder restore resolution and sharpen boundaries, and a final convolutional head maps features to image-space logits. This design keeps the implementation simple~(we adopt it largely as in SemiVL) while providing strong priors from text.
\vspace{0.5\baselineskip}

\subsubsection{Dense CLIP Distillation}
To stabilize semantics on unlabeled data, we distill from a frozen CLIP-based dense predictor~\(q(\cdot \mid u)\). This anchors unlabeled training to a non-drifting teacher. Drawing from MaskCLIP~\cite{zhou2022extract}, this zero-shot dense predictor extracts dense patch-level features from CLIP's image encoder, i.e., the value features of the last attention layer, computing similarities without fine-tuning the core model. Visual features are dotted with text embeddings to produce segmentation logits. Since the resulting predictions are highly noisy, we also apply confidence masking to guarantee precision. We adopt this to prevent noisy distillation, focusing on high-certainty regions. In our training logic, this is applied for auxiliary pipeline against two strong views outputs, using MaskCLIP pseudo-labels as targets for consistency:
\begin{equation}
\mathcal{L}_{\text{mclip-dual}} = \mathcal{L}_{\text{mclip}}(\alpha_{s1}, \alpha_{w}) + \mathcal{L}_{\text{mclip}}(\alpha_{s2}, \alpha_{w})  
\end{equation}

These distilled predictions provide weak supervision for the student, enabling self-training while keeping the model anchored to CLIP semantics during pretraining. 

In summary, we combine supervised learning on labeled source data, weak-to-strong consistency on unlabeled data, and MaskCLIP-based pseudo-label supervision on the two strong views:
\begin{equation}
\mathcal{L}_{\text{cent}}
= \mathcal{L}_{\text{sup}}
+ \lambda_{\text{cons}}\!\left(\mathcal{L}_{\text{cons}}^{(s_1)} + \mathcal{L}_{\text{cons}}^{(s_2)}\right)
+ \lambda_{\text{mclip}}\!\left(\mathcal{L}_{\text{mclip}}^{(s_1)} + \mathcal{L}_{\text{mclip}}^{(s_2)}\right)
\end{equation}

where the superscripts \(s_1,s_2\) denote the two strong augmentations \(\alpha_{s_1},\alpha_{s_2}\); each \(\mathcal{L}_{\text{mclip}}^{(s_j)}\) uses pseudo-labels from the frozen MaskCLIP teacher computed on the weak view \(\alpha_w(u)\) with confidence masking.
\subsection{Federated Learning}
In the federated learning stage, a central server coordinates with multiple clients to collaboratively train the model while preserving data privacy. The server begins by pretraining the model on the labeled source dataset using the centralized training approach described in the previous section. This pretrained model serves as the initial global model~$w^{(0)}$. In each communication round~$t$, the server broadcasts the current global model~$w^{(t)}$ to a subset of selected clients~$\mathcal{S}_t \subseteq \mathcal{K}$. Each participating client~$k \in \mathcal{S}_t$ initializes its local model with~$w^{(t)}$ and performs local training for~$E$~epochs on its private dataset~$\mathcal{D}_k$ using the surrogate objective~$\mathcal{L}_k$, which adapts the centralized components~(consistency regularization, Vision-Language decoder, and dense CLIP distillation) to the client's data. After local updates, clients send their updated models back to the server for aggregation.
\vspace{0.5\baselineskip}

\subsubsection{FedAvg for (semi-)supervised clients}
For the semi-supervised case, where clients have access to a small amount of labeled data, we employ the standard Federated~Averaging~(FedAvg) algorithm. In FedAvg, the server aggregates the client models as follows:
\begin{equation}
  w^{(t+1)} = \sum_{k \in \mathcal{S}_t} \frac{n_k}{\sum_{j \in \mathcal{S}_t} n_j} \, w_k^{(t)},
\end{equation}
where~$w_k^{(t)}$ is the updated model from client~$k$ after local training, and~$n_k$ is the size of client~$k$'s dataset. This weighted average ensures that clients with more data contribute proportionally more to the global model.
\vspace{0.5\baselineskip}

\subsubsection{FedSWA for unsupervised clients}
In the unsupervised case, where clients only have unlabeled data, our experimental results show that FedAvg can lead to unstable training and performance degradation due to the lack of direct supervision and potential drift in pseudo-labels. To address this, we adopt Federated~Stochastic~Weight~Averaging~(FedSWA)~\cite{liu2025fedswa}, which enhances generalization by incorporating a cyclical learning rate decay during local updates and exponential~moving~average~(EMA) aggregation at the server. In FedSWA, each client's local learning rate decays linearly from an initial value~$\eta_0$~to~$\delta \eta_0$ over~$N$ local iterations:
\begin{equation}
\eta_{i} = \eta_0 \left(1 - \frac{i}{N}\right) + \frac{i}{N} \delta \eta_0,
\end{equation}
where~$i$ is the local iteration index (from 0 to~$N-1$), $N$~is the total number of local iterations, and~$\delta \in~[0,1]$ controls the final learning rate factor. The local update rule is:
\begin{equation}
w_{k,j+1}^{(t)} = w_{k,j}^{(t)} - \eta_j \, \nabla \ell_k(w_{k,j}^{(t)}),
\end{equation}
where~$\nabla \ell_k$~is the mini-batch gradient approximation computed on client~$k$'s data, and~$j$ indexes the local steps. At the server, aggregation uses EMA:
\begin{equation}
w_{swa}^{(t)} = \frac{1}{|\mathcal{S}_t|} \sum_{k \in \mathcal{S}_t} w_k^{(t)}, \quad w^{(t+1)} = w^{(t)} + \gamma (w_{swa}^{(t)} - w^{(t)})
\label{eq:fedswa_server}
\end{equation}
where~$\gamma$ is the EMA coefficient, and~$w_{swa}^{(t)}$ is the simple average of the received client models.
Our implementation differs from the original FedSWA in two ways:~(i)~we set~$\alpha_{\text{EMA}}=1$, which reduces Eq.~\eqref{eq:fedswa_server}~to~$w^{(t+1)}=v_t$, and we replace the linear learning-rate decay with a
polynomial schedule,
\begin{equation}
  \eta_{t,e}
  \;=\;
  \eta_\ell \Big[ (1-s_e)^{p} \;+\; \rho\big(1-(1-s_e)^{p}\big) \Big],
  \qquad p \ge 1,
  \label{eq:poly_lr}
\end{equation}
for smoother scheduling.

\section{Experiments}

\subsection{Datasets and Partitioning}\label{sec:partitioning}
We evaluate our framework on two domain adaptation scenarios, each with a source and target dataset:
\subsubsection{Synthetic-to-Real:} The source is the GTA5~\cite{Richter_2016_ECCV}~dataset, a collection of 24,966 synthetic urban scene images rendered from the GTA5 game engine. Each GTA5 image is annotated for 19 semantic classes compatible with Cityscapes. The target is Cityscapes, a real-world street scene dataset with 5,000 high-resolution images (2,975 train, 500 val, 1,525 test) covering 19 classes. Cityscapes images are~$2048\times 1024$ pixels captured across 50 cities from Central Europe, providing diverse urban driving conditions.
\vspace{0.5\baselineskip}

\subsubsection{Clear-to-Adverse Weather:} The source is Cityscapes~(clear daytime images, as above), and the target is ACDC~(Adverse Conditions Dataset with Correspondences). ACDC consists of 4,006 finely-annotated images captured under adverse conditions~–~fog, nighttime, rain, and snow. The target label space in ACDC directly inherits the Cityscapes schema~(the standard 19 classes of road objects), making Cityscapes-to-ACDC a compatible domain adaptation setting. ACDC’s training set contains roughly 1,600 images~(about 400 per adverse condition), presenting a challenging domain shift from normal weather to each adverse weather type.
\vspace{0.5\baselineskip}

To adapt these setups to a federated learning scenario with heterogeneous client data, we partition the target datasets among many simulated clients. For Cityscapes~(used as target in the synthetic-to-real scenario), we adopt the heterogeneous split proposed by LADD~\cite{shenaj2023learning}: the 2,975 Cityscapes training images are divided into 144 clients, each client holding 10~to~45 images from a single city. This means each client’s data is biased to one city’s distribution, creating a non-i.i.d. setting as in real federated fleets. For ACDC~(target in the clear-to-adverse scenario), we build on the split from HyperFLAW~\cite{rizzoli2025cars} which originally defined 32 clients across three adverse conditions (Fog/Rain/Night from ACDC) plus a “clear-day” group from Cityscapes. In our case, to avoid using any Cityscapes “day” data during adaptation (preventing data leakage from the source pretraining), we remove the day-weather clients and instead include the snow condition~(which HyperFLAW had excluded). We thus obtain 28 clients in total, corresponding to the four adverse weather conditions~(fog, night, rain, snow), with each client containing 2–4 weather conditions.

\subsection{Source Model Pretraining}

Before federated adaptation, we obtain a strong initial model on the source data~(server-side pretraining). For the clear-to-adverse-weather scenario~(Cityscapes~$\to$~ACDC), we start from a SemiVL checkpoint trained on Cityscapes with half of the images labeled~(and half treated as unlabeled)~to incorporate Vision-Language guidance. The model uses a ViT-B/16 backbone initialized from CLIP~\cite{radford2021learning}. We then fine-tune this model on the full Cityscapes training set~(all 2,975 images) using UniMatch V2 training logic. The fine-tuning is run for a total of 360 epochs, which, given our smaller per-GPU batch size, yields an equivalent number of update iterations as the original SemiVL schedule of 240 epochs at batch size 8. We use a batch size of 2 per GPU~(effective batch~=~4 with two H100 GPUs), and optimize with AdamW. The initial learning rate is~$5\times 10^{-5}$, decayed with a polynomial schedule~(power 0.9) over epochs. Following SemiVL, input images are randomly cropped to~$801\times 801$~pixels for training. We apply strong data augmentations including random color jittering, grayscale, CutMix, random scaling, and random crops to improve generalization. We also employ sliding-window inference at test time as used in UniMatch. During training, we set the confidence threshold~$\tau = 0.9$ for pseudo-label filtering. Moreover, our dense CLIP guidance loss weight~($\lambda_{mclip}$)~starting at 0.1 and linearly decaying to 0 by the end of training. This CLIP distillation encourages the model’s features to align with text-defined class prototypes early in training, and then gradually lets the model focus more on its own consistency signals. We adopt the same text prompt engineering for class definitions as SemiVL. 
\vspace{0.5\baselineskip}

For the synthetic-to-real scenario~(GTA5~$\to$~Cityscapes), we pretrain a model on the GTA5 source data from scratch. We train the model on the 23,905 GTA5 images with supervised segmentation loss. The training schedule is 120 epochs with an initial learning rate of~$5\times 10^{-5}$~(poly decay 0.9). We use a total batch size of 9 images for GTA5 training. The input resolution and augmentations are kept the same as above:~$801\times 801$~random crops, with color jitter, grayscale conversion, CutMix, and random cropping applied. We also set~$\tau = 0.9$ for pseudo-label confidence.

\subsection{Benchmark Protocol}

We evaluate our approach under several training/test settings to compare against baselines:

\subsubsection{Source Only}
We first examine the source-trained model’s performance on the target domain without any adaptation. This corresponds to directly testing the pretrained source model on the target dataset~(e.g. Cityscapes-trained model on ACDC, or GTA5-trained model on Cityscapes) with no further training. It serves as a lower-bound reference, indicating how well the model generalizes to the new domain before adaptation. This also allows comparison to any prior works that evaluate zero-shot domain adaptation/generalization from Cityscapes to ACDC or GTA5 to Cityscapes.
\vspace{0.5\baselineskip}

\subsubsection{Target-Supervised Fine-Tuning (TSFT)} 
As an upper-bound baseline, we consider an approach where the source model is fine-tuned on labeled target data~(this protocol is defined in LADD). We denote TSFT-F for fine-tuning on the full target training set with ground truth, and TSFT-1/4 for fine-tuning on only 25\% of target labels. For example, in the ACDC case, TSFT-1/4 means the model is fine-tuned using 1/4 of ACDC’s annotated images. Concretely, we select 100 labeled images from each adverse condition in ACDC~(100 fog + 100 night + 100 rain + 100 snow = 400 total, which is 1/4 of the~$\sim$1600 train images) to simulate a low-label regime. TSFT-F uses all available target train labels~(e.g. all 1600 ACDC train images). These fine-tuning baselines represent what a federated client could achieve if a certain amount of target ground truth were available. We include them primarily in the heterogeneous ACDC experiments to highlight the gap between fully supervised adaptation and our semi-supervised federated approach. It is worth noting that TSFT can be implemented in two settings, namely centralized and federated.
\vspace{0.5\baselineskip}

\subsubsection{Centralized Unsupervised Self-Training~(CUST)} 
This baseline evaluates a traditional non-federated unsupervised domain adaptation by self-training. We take the source-pretrained model and perform self-training on the target data in a centralized setting. The model is trained on target without labels, using pseudo-labels and consistency training~(the same algorithm as our method uses on clients, but running centrally). For effective distillation, we set the CLIP guidance weight~$\lambda_{mclip} = 0.1$, initial LR~$5\times 10^{-5}$~(poly decay), crop size 801. This CUST baseline indicates the performance of self-training if we ignore federated constraints. The gap between CUST and our federated method shows the impact of data decentralization and client heterogeneity.
\vspace{0.5\baselineskip}

Finally, for the GTA5$\to$Cityscapes benchmarks, in addition to the aforementioned baselines, we also compare to results from LADD\cite{shenaj2023learning}, who introduced the federated heterogeneous Cityscapes split. We include LADD’s reported numbers for baselines like source-only, fine-tuning, and their own federated method to contextualize our performance on the same split.

\subsection{Federated Learning Setup}
We deploy a standard Federated~Stochastic~Weight~Averaging~(FedSWA) algorithm for unsupervised domain adaptation on the distributed clients. In each communication round, a subset of clients trains locally on their unlabeled data and the server aggregates their model updates. We tailor the federation hyperparameters for each scenario as follows:

\subsubsection{Cityscapes~$\to$~ACDC~(Clear-to-Adverse):} We simulate a federation of 28 clients~(as~described~\ref{sec:partitioning}). At each global round, we randomly sample 5 clients~($\approx 18\%$ of clients) to participate and perform one epoch of local training on their own data. We run 200 communication rounds in total, which ensures each client gets involved multiple times. The local optimizer on each client is AdamW with an initial LR of~$5\times 10^{-5}$, and we apply a polynomial learning rate decay across the 200 rounds. Each client’s local epoch uses a batch size of 2 images.
\vspace{0.5\baselineskip}

\subsubsection{GTA5~$\to$~Cityscapes~(Synthetic-to-Real):} we have 144 clients~(Cityscapes split) participating. We again select 5 clients per round for training. Due to the greater number of total clients and the highly fragmented data~(some clients have as few as 10 images), we increase the local training iterations: each selected client performs 10 local epochs on its data in a round. Following LADD, we run 300 rounds of communication. The optimizer and learning rate schedule are the same as above:~AdamW with initial~$5\times 10^{-5}$ and polynomial decay over 300 rounds. We use batch size 2 for local training as well.
\section{Experimental Results}
This section presents an in-depth analysis of a novel framework for Federated, source-Free, domain Generalization (FFREEDG), designed to address the significant domain shift from clear weather conditions (Cityscapes dataset) to adverse conditions (Adverse Conditions Dataset, ACDC) and from synthetic (GTA5) to real (Cityscapes). The core challenge lies in training a robust model in a decentralized setting without access to the original source data or any target-side ground-truth labels. The experimental results, detailed in Table~\ref{tab:city2acdc_results}, demonstrate that the proposed FFREEDG method on Cityscapes~$\to$~ACDC successfully mitigates this domain gap and achieves highly competitive performance against both SOTA domain generalization~(DG) and domain adaptation~(DA) methods. A critical finding is that the choice of aggregation strategy is paramount for stability in unsupervised federated learning, with a stochastic weight averaging approach proving more robust than traditional methods.

\subsection{Cityscapes $\to$ ACDC}
To contextualize the performance of the proposed methods, a series of standardized benchmark protocols were employed. The results, as summarized in Table~\ref{tab:city2acdc_results}, provide a comprehensive overview of the framework's effectiveness across different settings.

\begin{table*}[t]
\centering
\caption{Experimental results on Cityscapes$\to$ACDC (mIoU, \%). ACDC training images are partitioned into 28 clients. DAFormer~\cite{hoyer2022daformer}, SePiCo~\cite{xie2023sepico}, HRDA~\cite{hoyer2022hrda}, VLTSeg~\cite{hummer2023vltseg}, STA~\cite{gong2023train}, and PromptFormer~\cite{gong2023prompting} are domain adaptation/generalization baselines used for comparison with our source-only setting. \textbf{Abbreviations:} mIoU = mean Intersection over Union; DA = domain adaptation; DG = domain generalization; TSFT = target-supervised fine-tuning; TSFT-F = TSFT using all target labels; TSFT-1/4 = TSFT using 25\% of target labels; CUST = centralized unsupervised self-training; FFREEDG = Federated source-Free Domain Generalization.}
\label{tab:city2acdc_results}
\setlength{\tabcolsep}{6pt}
\renewcommand{\arraystretch}{1.12}
\resizebox{\textwidth}{!}{
\begin{tabular}{l c *{4}{c} !{\vrule width 0.8pt} *{3}{c}}
\toprule
\multirow{3.5}{*}{Setting} & \multicolumn{8}{c}{Method} \\
\cmidrule(lr){2-9}
& \multirow{2.5}{*}{FRIEREN (ours)} & \multicolumn{4}{c}{DA} & \multicolumn{3}{c}{DG} \\
\cmidrule(lr){3-6}\cmidrule(lr){7-9}
& & DAFormer~(1024$\times$512)~\cite{hoyer2022daformer} & SePiCo~(640$\times$640)~\cite{xie2023sepico} & PromptFormer~(1024$\times$512)~\cite{gong2023prompting} & HRDA~(512$\times$512)~\cite{hoyer2022hrda}
& VLTSeg (512$\times$512)~\cite{hummer2023vltseg} & STA~(1024$\times$512)~\cite{gong2023train} & VLTSeg (1024$\times$1024)~\cite{hummer2023vltseg} \\
\midrule
Cityscapes           & 81.86 & --   & --   & --   & --    & --   & --   & --   \\
Source Only          & 61.85 & --   & --   & --   & --    & 59.84 & 60.9 & 77.91 \\
Centralized TSFT     & 77.89 & --   & --   & --   & --    & --   & --   & --   \\
CUST                 & 66.08 & --   & --   & --   & --    & --   & --   & --   \\
Federated TSFT-F     & 75.43 & --   & --   & --   & --    & --   & --   & --   \\
Federated TSFT-1/4   & 73.58 & --   & --   & --   & --    & --   & --   & --   \\
FFREEDG              & 64.43 & 55.4 & 59.1 & 62.0 & 68.0  & --   & --   & --   \\
\bottomrule
\end{tabular}}
\end{table*}

\paragraph{Centralized Learning}
A fundamental challenge in this setup is the domain shift between the pristine urban scenes of Cityscapes and the challenging adverse weather conditions of ACDC. The pretrained model, which achieves a strong mIoU~of~81.86\% on Cityscapes, suffers a substantial performance degradation when tested in a zero-shot manner on ACDC, yielding an mIoU of only~61.85\%. This considerable drop of~20.01\% provides empirical evidence of the severe domain gap, which is the central problem that the proposed framework aims to address. The source-only result is therefore not a measure of the method's robustness, but rather a direct measure of the problem's magnitude.
\vspace{0.5\baselineskip}

To validate the capabilities of the core training methodology, which includes self-training, weak-to-strong consistency regularization, and dense CLIP distillation—a centralized unsupervised self-training~(CUST) experiment was conducted. The CUST method, which applies the proposed training objective to the unlabeled ACDC dataset in a centralized manner, achieves an mIoU of~66.08\%. This result represents a significant gain of~+4.23\% over the source-only baseline, confirming that the underlying training logic is effective at adapting to the target domain, even without ground-truth supervision. This validates the effectiveness of using consistency signals and distillation from a vision foundation model to refine the semantic representations of the model on unlabeled data.
\vspace{0.5\baselineskip}

\paragraph{Federated Learning}
\method{} achieves a final mIoU of~64.43\% under FFREEDG. Relative to the source-only baseline~(61.85\%), this represents a~+2.58\% absolute improvement. CUST achieves~66.08\%, outperforming both FFREEDG and source-only by~1.65\% and~4.23\%, respectively. This finding indicates that the framework successfully transfers the effectiveness of the centralized approach to a decentralized setting while incurring a minimal performance penalty for the privacy-preserving, source-free, and non-IID data constraints of the federated environment.
\vspace{0.5\baselineskip}

The performance of our method is highly competitive when compared to other SOTA methods. The framework outperforms a number of leading DG methods, including PromptFormer~(62.0\%) and STA~(60.9\%). Notably, the method also surpasses the performance of VLTSeg~(512$\times$512)~by~+4.53\%, a strong contemporary DG model operating at a comparable resolution. In comparison to top DA methods, our framework remains highly competitive. The performance is only~$3.57$\% lower than HRDA~(68.0\%), a prominent DA method. While VLTSeg~(1024$\times$1024) sets a new SOTA benchmark with an mIoU of~77.91\%, its superior performance can be attributed to the use of a larger backbone, an advanced decoder~(Mask2Former~\cite{cheng2022masked}), and higher-resolution training, which presents a clear avenue for future research to further enhance our framework's capabilities.
\vspace{0.5\baselineskip}

\paragraph{Unsupervised Federated Learning}
A key observation from the experimental results is the sensitivity of unsupervised federated learning to the model aggregation strategy. Initial experiments using FedAvg showed that the performance of the model could degrade over the course of training. This instability is a direct consequence of the unsupervised setting: because there are no ground-truth labels on the clients, local model updates are based on potentially noisy and “drifting” pseudo-labels. In this context, FedAvg, which aggregates client updates in proportion to their dataset size, can be susceptible to a large client with noisy pseudo-labels disproportionately influencing and destabilizing the global model. This is a fundamental limitation of applying an aggregation method designed for a supervised, labeled setting to a problem where direct supervision is absent. To address this challenge, our framework adopts Federated~Stochastic~Weight~Averaging~(FedSWA) for unsupervised clients. FedSWA mitigates these stability issues via cyclical learning-rate decay and Stochastic~Weight~Averaging. The successful performance of our method is a testament to the importance of a carefully selected aggregation strategy that is purpose-built to address the unique challenges of unsupervised federated learning.
\vspace{0.5\baselineskip}

\paragraph{Semi-Supervised and Supervised Benchmarks.}
Further analysis of the fine-tuning baselines provides additional context for the challenges of federated learning. The TSFT-F protocol, which fine-tunes the source-trained model on the full, labeled target dataset, achieves an mIoU of~77.89\% in a centralized setting. When this same protocol is applied in a federated manner with labeled data distributed among clients, the performance drops to~75.43\%. This performance gap of~2.46\% between centralized and federated supervised fine-tuning is a compelling finding. It demonstrates that the performance trade-off associated with data decentralization is not unique to our unsupervised approach but is an inherent challenge of federated learning itself. This validates our observation that the~1.65\% difference between CUST and FFREEDG is an inevitable cost for the benefits of a decentralized framework.
\vspace{0.5\baselineskip}

The TSFT-1/4 benchmark, which fine-tunes on only~25\% of the target's labeled data, provides a powerful argument for the efficiency of semi-supervised approaches. This method, which requires only 400 labeled images from the ACDC dataset—a significant reduction in annotation effort—achieves an mIoU of~73.58\%. This result is highly competitive, demonstrating that with a small amount of labeled data, performance can approach that of full supervision~(TSFT-F, 75.43\%~mIoU) while substantially reducing the cost of data annotation. This benchmark highlights the potential for methods that leverage limited labeled data. In contrast, our proposed unsupervised method achieves a strong mIoU~of~64.43\% using zero ground-truth labels from the target domain, which demonstrates the immense value of our framework as a practical and privacy-preserving solution for domain generalization when labeled data is completely unavailable.
\subsection{GTA5~$\to$~Cityscapes}
Following the analysis on the Cityscapes~$\to$~ACDC domain, we extend our evaluation to an even more challenging scenario: transferring from the synthetic GTA5 dataset to the real-world Cityscapes dataset. This domain shift is particularly significant due to the inherent visual differences between computer-generated and real-world imagery. The experimental results, as detailed in Table~\ref{tab:gta2city_results}, were conducted on the highly heterogeneous dataset proposed by LADD, which fragments the Cityscapes training set into 144 clients.

\begin{table}[t]
\centering
\caption{Results on GTA5~$\to$~Cityscapes~(mIoU, \%). Cityscapes train is partitioned into 144 clients following LADD.}
\label{tab:gta2city_results}
\setlength{\tabcolsep}{6pt}
\renewcommand{\arraystretch}{1.12}
\resizebox{\linewidth}{!}{
\begin{tabular}{lcccc}
\toprule
\multirow{3.5}{*}{Setting} & \multicolumn{4}{c}{Method} \\
\cmidrule(lr){2-5}
& \multirow{2.5}{*}{FRIEREN (ours)} & \multicolumn{3}{c}{UDA} \\
\cmidrule(lr){3-5}
& & LADD~\cite{shenaj2023learning} & MCD~\cite{saito2018maximum} & DAFormer~\cite{hoyer2022daformer} \\
\midrule
Source Only        & 29.72 & 24.05 & 20.55 & 42.31 \\
Centralized TSFT   & 81.86 & 66.64 & --    & --    \\
Federated TSFT-F   & 55.99 & 59.40 & --    & --    \\
CUST               & 31.48 & --    & --    & --    \\
FFREEDG            & 31.35 & 36.49 & 10.86 & --    \\
\bottomrule
\end{tabular}}
\end{table}
\vspace{0.5\baselineskip}

\paragraph{Source-Only Performance}
The severity of the domain gap is evident from the source-only performance. Our model pretrained on GTA5 achieves an mIoU of~29.72\% when tested directly on Cityscapes. Notably, our source-only model shows a substantial lead over the LADD baseline’s source-only result~(24.05\%), indicating that our initial pretraining yields a more robust and better-generalized starting point.
\vspace{0.5\baselineskip}

\paragraph{Federated and Centralized Baselines}
\method\ attains 31.35\% mIoU on this challenging dataset. While this may appear modest at first glance, a closer comparison to established baselines reveals its strength: our method significantly outperforms both the centralized MCD~baseline~(20.55\%) and the federated UDA variant of MCD~(10.86\%). This demonstrates that our framework effectively leverages unsupervised learning in a federated environment, surpassing traditional methods that struggle under the same constraints.
\vspace{0.5\baselineskip}

When compared with LADD’s FFREEDA method~(36.49\%), it is important to note the fundamental difference in problem settings. LADD is a domain adaptation~(DA) approach using FDA~\cite{yang2020fda}, designed to perform well on the specific target domain, whereas our FFREEDG is a domain generalization~(DG) method aiming to generalize to unseen distributions by training across multiple domains—an inherently more challenging objective. The marginal gap between our FFREEDG and LADD’s FFREEDA underscores our method’s viability as a generalizable solution. In a supervised fine-tuning setting with target labels, our TSFT-F result of~55.99\% is lower than LADD’s corresponding TSFT result~(59.4\%), which is a reasonable and expected trade-off given that LADD is tailored for DA.
\vspace{0.5\baselineskip}

\section{Ablation Studies}
\subsection{Federated Aggregation}
The FedSWA research paper also proposes FedMoSWA~\cite{liu2025fedswa}, which adds momentum to the client updates and control variates for variance reduction. We quantitatively evaluated FedMoSWA and compared it to FedAvg, finding that it does not improve performance and instead performs worse than FedAvg~(see~Table~\ref{tab:fedswa_ablation}). Furthermore, prior work has similarly observed that adding FedProx~\cite{li2020federated} or SCAFFOLD-style~\cite{karimireddy2020scaffold} adjustments does not help in dense prediction tasks such as semantic segmentation~\cite{shenaj2023learning,miao2023fedseg}; FedSeg~\cite{miao2023fedseg} report that under non-IID data, FedProx~\cite{li2020federated}, FedDyn~\cite{jin2023feddyn} and MOON~\cite{li2021model} achieve segmentation performance similar to or even lower than FedAvg, indicating that coarse regularization of entire model weights or representations is ineffective for dense prediction. Therefore, we adopt the simpler FedSWA approach in our unsupervised federated training to stabilize learning.
\begin{table}[t]
\centering
\caption{Federated aggregation algorithm comparison on Cityscapes$\to$ACDC (mIoU).}
\label{tab:fedswa_ablation}
\resizebox{0.35\columnwidth}{!}{
\begin{tabular}{lc}
\toprule
\textbf{Method} & \textbf{mIoU} \\
\midrule
Source Only & 20.97 \\
FedAvg      & 36.91 \\
FedMoSWA    & 27.66 \\
\bottomrule
\end{tabular}
}
\end{table}
\begin{table}[t]
\centering
\caption{Fine-tuning strategy comparison on GTA5$\to$Cityscapes.
Models are pretrained on GTA5 at $512\times512$ for 100 epochs.}
\label{tab:peft_gta_cs}
\setlength{\tabcolsep}{4pt}
\renewcommand{\arraystretch}{1.05}
\footnotesize
\resizebox{0.7\columnwidth}{!}{%
\begin{tabular}{lcc}
\toprule
\textbf{Fine-tuning Method} & \makecell{\textbf{Updated Params}\\\textbf{(Enc.)}}
& \makecell{\textbf{mIoU}} \\
\midrule
Rein~\cite{wei2024stronger}              & 1.7M  & 24.12 \\
Spatial Fine-Tuning~\cite{hoyer2024semivl} & 29.1M & 23.85 \\
\bottomrule
\end{tabular}%
}
\end{table}

\subsection{Fine-Tuning Strategies}
Table~\ref{tab:peft_gta_cs} compares a parameter-efficient fine-tuning (PEFT) method, Rein~\cite{wei2024stronger}, with the Spatial Fine-Tuning strategy of SemiVL. Our model generalizes poorly on GTA5$\to$Cityscapes. In contrast, Rein has been reported to generalize strongly on this benchmark and to outperform LoRA~\cite{hu2022lora}. In our setting it attains 24.12 mIoU, slightly higher than Spatial Fine-Tuning (23.85 mIoU), while requiring fewer trainable parameters. However, Rein’s additional adapter modules increase the memory footprint and implementation complexity, and the gain is marginal. Therefore, for simplicity and stability, we have adopted Spatial Fine-Tuning in the experiments of this work.

\paragraph{Future Work}
This analysis confirms the effectiveness and robustness of the proposed FFREEDG framework for source-free federated domain generalization. The results demonstrate that the framework successfully reduces the domain gap from Cityscapes to ACDC, achieving a final mIoU of~64.43\%, which is highly competitive with SOTA methods. A key contribution is the empirical validation of using a combination of self-training, consistency regularization, and distillation from vision foundation models to adapt effectively to unseen domains without target-side supervision. Furthermore, the analysis provides critical insights into the nature of federated learning for this task:
\begin{enumerate}
\item The performance trade-off between centralized and decentralized training is a general characteristic of the federated setting, not a specific limitation of our unsupervised method.
\item The choice of aggregation strategy is of paramount importance in unsupervised federated learning, with methods like FedSWA offering superior stability and robustness by directly addressing pseudo-label drift.
\end{enumerate}

Although our method may not achieve SOTA performance against specialized DG or DA methods on all benchmarks, it demonstrates the feasibility of continuous learning in unsupervised and semi-supervised modes within a federated setting, especially valuable where labeled data is expensive and scarce. Our method is a foundational learning strategy that can be extended and integrated with other DG or DA methods to boost performance. The modular components in our model can be replaced with more advanced ones; for example, our current ViT-B/16 distillation objective could be upgraded to stronger foundation models (e.g., DINOv2, DINOv3~\cite{simeoni2025dinov3}) to further increase robustness to domain shift. Future work may integrate higher-resolution backbones and larger-scale vision foundation models to close the gap with SOTA methods. It may also incorporate alternative decoders (e.g., Mask2Former~\cite{cheng2022masked}) to push the boundaries of federated semantic segmentation.

\section{Conclusion}
In this work we introduced \textbf{FFREEDG}, a realistic and challenging setting for~\emph{source-free federated domain generalization} in semantic segmentation. We proposed a vision–language guided framework, \textbf{\method}, to address this setting, employing (i) UniMatch-style weak-to-strong consistency with self-training and dense CLIP distillation to regularize learning on unlabeled data, and (ii) a stabilization strategy based on FedSWA for unsupervised aggregation in the presence of noisy pseudo-labels. Our approach attains competitive performance under strict privacy constraints. On~Cityscapes$\to$ACDC, FFREEDG reaches~64.43\% mIoU, closing most of the gap to centralized self-training~(66.08\%) and outperforming recent DG baselines while remaining close to strong DA methods. On the heterogeneous~GTA5$\to$Cityscapes~split~(144 clients), our source-only~(29.72\%) and FFREEDG~(31.35\%) exceed classical UDA baselines. This analysis confirms the effectiveness and robustness of the proposed FFREEDG framework and the results demonstrate that the framework successfully transfers the effectiveness of the centralized approach to a decentralized setting while incurring a minimal performance penalty for the privacy-preserving, source-free, and non-IID data constraints of the federated environment, which is a reference for future research on unsupervised semantic segmentation in federated learning.
\section*{Acknowledgment}
The author is deeply grateful to Associate Professor Dr. Ir. Fons van der Sommen and Dr. Ir. Christiaan Viviers and Dr. Ir. Willem Sanberg and Dr. Fabrizio Piva for their insightful feedback and guidance. This project was carried out using the Snellius supercomputer, with computational resources provided by the Dutch~Research~Council~(NWO).

\end{document}